\documentclass[11pt]{article}
\usepackage{acl2016}
\usepackage{times}
\usepackage{url}
\usepackage{latexsym}
\usepackage{graphicx}
\usepackage{subfig}
\usepackage{amssymb}
\usepackage{amsmath}
\usepackage{amsfonts}
\usepackage{pgfplots}
\usepackage{kotex}
\usepackage{CJKutf8}
\usepackage{multirow}
\usepackage{enumitem}
\usepackage{arydshln}
\usepackage{soul}
\setlist{itemsep=0pt}

\aclfinalcopy 

\title{SYSTRAN's Pure Neural Machine Translation Systems}

            \author{Josep Crego, Jungi Kim, Guillaume Klein, Anabel Rebollo, Kathy Yang, Jean Senellart\\
{\normalsize\textbf{Egor Akhanov, Patrice Brunelle, Aur\'elien Coquard, Yongchao Deng, Satoshi Enoue, Chiyo Geiss}} \\
{\normalsize\textbf{Joshua Johanson, Ardas Khalsa, Raoum Khiari, Byeongil Ko, Catherine Kobus, Jean Lorieux}} \\
{\normalsize\textbf{Leidiana Martins, Dang-Chuan Nguyen, Alexandra Priori, Thomas Riccardi, Natalia Segal}} \\
{\normalsize\textbf{Christophe Servan, Cyril Tiquet, Bo Wang, Jin Yang, Dakun Zhang, Jing Zhou, Peter Zoldan}}\\ \\
  SYSTRAN \\
  {\tt firstname.lastname@systrangroup.com} \\}

\date{}

\begin{document}
\maketitle

\begin{abstract}
\label{abstract}

Since the first online demonstration of Neural Machine Translation (NMT) by LISA \cite{DBLP:journals/corr/BahdanauCB14}, NMT development has recently moved from laboratory to production systems as demonstrated by several entities announcing roll-out of NMT engines to replace their existing technologies. NMT systems have a large number of training configurations and the training process of such systems is usually very long, often a few weeks, so role of experimentation is critical and important to share. In this work, we present our approach to production-ready systems simultaneously with release of online demonstrators covering a large variety of languages ($12$ languages, for $32$ language pairs). We explore different practical choices: an efficient and evolutive open-source framework; data preparation; network architecture; additional implemented features; tuning for production; {\it etc}. We discuss about evaluation methodology, present our first findings and we finally outline further work.

Our ultimate goal is to share our expertise to build competitive production systems for "{\it generic}" translation. We aim at contributing to set up a collaborative framework to speed-up adoption of the technology, foster further research efforts and enable the delivery and adoption to/by industry of use-case specific engines integrated in real production workflows. Mastering of the technology would allow us to build translation engines suited for particular needs, outperforming current simplest/uniform systems.


\end{abstract}

\section{Introduction}
\label{sec:intro}

Neural MT has recently achieved state-of-the-art performance in several large-scale translation tasks. As a result, the deep learning approach to MT has received exponential attention, not only by the MT research community but by a growing number of private entities, that begin to include NMT engines in their production systems.

In the last decade, several open-source MT toolkits have emerged---Moses \cite{Koehn07} is probably the best-known out-of-the-box MT system---coexisting with commercial alternatives, though lowering the entry barriers and bringing new opportunities on both research and business areas. Following this direction, our NMT system is based on the open-source project \texttt{seq2seq-attn}\footnote{\url{https://github.com/harvardnlp/seq2seq-attn}} initiated by the Harvard NLP group\footnote{\url{http://nlp.seas.harvard.edu}} with the main contributor Yoon Kim.
We are contributing to the project by sharing several features described in this technical report, which are available to the MT community.

Neural MT systems have the ability to directly model, in an end-to-end fashion, the association from an input text (in a source language) to its translation counterpart (in a target language). A major strength of Neural MT lies in that all the necessary knowledge, such as syntactic and semantic information, is learned by taking the global sentence context into consideration when modeling translation. However, Neural MT engines are known to be computationally very expensive, sometimes needing for several weeks to accomplish the training phase, even making use of cutting-edge hardware to accelerate computations. Since our interest is for a large variety of languages, and that based on our long experience with machine translation, we do not believe that a one-fits-all approach would work optimally for languages as different as Korean, Arabic, Spanish or Russian, we did run hundreds of experiments, and particularily explored language specific behaviors. One of our goal would indeed be to be able to inject existing language knowledge in the training process.

In this work we share our recipes and experience to build our first generation of production-ready systems for ``generic'' translation, setting a starting point to build specialized systems. We also report on extending the baseline NMT engine with several features that in some cases increase performance accuracy and/or efficiency while for some others are boosting the learning curve, and/or model speed. As a machine translation company, in addition to decoding accuracy for ``generic domain'', we also pay special attention to features such as:

\begin{itemize}\itemsep0em
\item Training time
\item Customization possibility: user terminology, domain adaptation
\item Preserving and leveraging internal format tags and misc placeholders
\item Practical integration in business applications: for instance online translation box, but also translation batch utilities, post-editing environment...
\item Multiple deployment environments: cloud-based, customer-hosted environment or embedded for mobile applications
\item {\it etc}
\end{itemize}

More important than unique and uniform translation options, or reaching state-of-the-art research systems, our focus is to reveal language specific settings, and practical tricks to deliver this technology to the largest number of users.

The remaining of this report is as follows: Section \ref{sec:systemDescription} covers basic details of the NMT system employed in this work. Description of the translation resources are given in section \ref{sec:corpora}. We report on the different experiments for trying to improve the system by guiding the training process in section \ref{sec:technology} and section \ref{sec:performance}, we discuss about performance. In section \ref{sec:evaluation} and \ref{sec:practical}, we report on evaluation of the models and on practical findings. And we finish by describing work in progress for the next release.

\section{System Description}
\label{sec:systemDescription}

We base our NMT system on the encoder-decoder framework made available by the open-source project \texttt{seq2seq-attn}.
With its root on a number of established open-source projects such as Andrej Karpathy's char-rnn,\footnote{\url{https://github.com/karpathy/char-rnn}}
Wojciech Zaremba's standard long short-term memory (LSTM)\footnote{\url{https://github.com/wojzaremba/lstm}}
and the rnn library from Element-Research,\footnote{\url{https://github.com/Element-Research/rnn}}
the framework provides a solid NMT basis consisting of LSTM, as the recurrent module and
faithful reimplementations of \emph{global-general-attention} model and \emph{input-feeding} at each time-step of the RNN decoder
as described by \newcite{luong-pham-manning:2015:EMNLP}.

It also comes with a variety of features such as the ability to train with bidirectional encoders and pre-trained word embeddings, the ability to handle unknown words during decoding by substituting them either by copying the source word with the most attention or by looking up the source word on an external dictionary, and the ability to switch between CPU and GPU for both training and decoding.
The project is actively maintained by the Harvard NLP group\footnote{\url{http://nlp.seas.harvard.edu}}.

Over the course of the development of our own NMT system, we have implemented additional features as described in Section \ref{sec:technology}, and contributed back to the open-source community by making many of them available in the \texttt{seq2seq-attn} repository.

\texttt{seq2seq-attn} is implemented on top of the popular scientific computing library \emph{Torch}.\footnote{\url{http://torch.ch}}
Torch uses \emph{Lua}, a powerful and light-weight script language, as its front-end and uses the \emph{C} language where efficient implementations are needed.
The combination results in a fast and efficient system both at the development and the run time.
As an extension, to fully benefit from multi-threading, optimize CPU and GPU interactions, and to have finer control on the objects for runtime (sparse matrix, quantized tensor, ...), we developed a \emph{C}-based decoder using the \emph{C} APIs of \emph{Torch}, called C-torch, explained in detail in section \ref{ssec:cdecoder}.




The number of parameters within an NMT model can grow to hundreds of millions,
but there are also a handful of meta-parameters that need to be manually determined.
For some of the meta-parameters, many previous work presents clear choices on their effectiveness, such as using the attention mechanism or feeding the previous prediction as input to the current time step in the decoder.
However, there are still many more meta-parameters that have different optimal values across datasets, language pairs, and the configurations of the rest of the meta-parameters.
In table \ref{tab:metaparameters}, we list the meta-parameter space that we explored during the training of our NMT systems. 

In appendix \ref{sec:parameters}, we detail the parameters used for the online system of this first release.


\begin{table}
\begin{center}
\begin{tabular}{l l}
Model & \\
\hline
&	Embedding dimension: 400-1000\\
&	Hidden layer dimension: 300-1000\\
&	Number of layers: 2-4\\
&	Uni-/Bi-directional encoder \\
\hline

Training & \\
\hline
&	Optimization method \\
&		Learning rate \\
&		Decay rate \\
&		Epoch to start decay \\
&	Number of Epochs \\
&	Dropout: 0.2-0.3\\
\hline
Text unit & Section \ref{ssec:tokenization} \\
\hline
& Vocabulary selection \\
& Word vs. Subword (e.g. BPE) \\
\hline
Train data & Section \ref{sec:corpora} \\
\hline
&	size (quantity vs. quality) \\
&	max sentence length \\
&	selection and mixture of domains \\
\end{tabular}
\end{center}
\caption{\label{tab:metaparameters} There are a large number of meta-parameters to be considered during training. The optimal set of configurations differ from language pair to language pair.}
\end{table}

\section{Training Resources}
\label{sec:corpora}

Training ``generic'' engines is a challenge, because there is no such notion of generic translation which is what online translation service users are expecting from these services. Indeed online translation is covering a very large variety of use cases, genres and domains. Also available open-source corpora are domain specific: Europarl \cite{koehn2005europarl}, JRC \cite{steinberger2006jrc} or MultiUN \cite{chen2012multiun} are legal texts, ted talk are scientific presentations, open subtitles \cite{tiedemann2012parallel} are colloquial, {\it etc.} 
As a result, the training corpora we used for this release were built by doing a weighted mix all of the available sources. For languages with large resources, we did reduce the ratio of the institutional (Europal, UN-type), and colloquial types -- giving the preference to news-type, mix of webpages (like Gigaword).

Our strategy, in order to enable more experiments was to define 3 sizes of corpora for each language pair: a baseline corpus (1 million sentences) for quick experiments (day-scale), a medium corpus (2-5M) for real-scale system (week-scale) and a very large corpora with more than 10M segments.

The amount of data used to train online systems are reported in table \ref{tab:corpora}, while most of the individual experimental results reported in this report are obtained with baseline corpora.

\begin{table*}[h]
\small
\begin{center}
\begin{tabular}{|c||c|cc|cc||c|cc|cc||cc|}

\hline  \multirow{3}{*}{\shortstack{Language\\ Pair}} & \multicolumn{5}{|c||}{Training} & \multicolumn{7}{|c|}{Testing} \\
\cline{2-13}  & \#Sents & \multicolumn{2}{|c|}{\#Tokens}  & \multicolumn{2}{|c||}{Vocab} & \#Sents & \multicolumn{2}{|c|}{\#Tokens}  & \multicolumn{2}{|c||}{Vocab} & \multicolumn{2}{|c|}{OOV} \\
              &                  & source & target                 &  source & target             &         &  source & target                &  source & target             & source & target \\
\hline
en$\leftrightarrow$br & 2.7M & 74.0M & 76.6M & 150k & 213k & 2k & 51k & 53k & 6.7k & 8.1k & 47 & 64  \\
en$\leftrightarrow$it & 3.6M & 98.3M & 100M & 225k & 312k & 2k & 52k & 53k & 7.3k & 8.8k & 66 & 85 \\
en$\leftrightarrow$ar & 5.0M & 126M & 155M & 295k & 357k & 2k & 51k & 62k & 7.5k & 8.7k & 43 & 47 \\
en$\leftrightarrow$es & 3.5M & 98.8M & 105M & 375k & 487k & 2k & 53k & 56k & 8.5k & 9.8k & 110 & 119 \\
en$\leftrightarrow$de & 2.6M & 72.0M & 69.1M & 150k & 279k & 2k & 53k & 51k & 7.0k & 9.6k & 30 & 77 \\
en$\leftrightarrow$nl & 2.1M & 57.3M & 57.4M & 145k & 325k & 2k & 52k & 53k & 6.7k & 7.9k & 50 & 141 \\
en$\rightarrow$ko & 3.5M & 57.5M & 46.4M & 98.9k & 58.4k & 2k & 30k & 26k & 7.1k & 11k & 0 & - \\
\hline
en$\leftrightarrow$fr & 9.3M  & 220M & 250M & 558k & 633k & 2k & 48k & 55k & 8.2k & 8.6k & 77 & 63    \\
\hline
fr$\leftrightarrow$br & 1.6M & 53.1M & 47.9M & 112k & 135k & 2k & 62k & 56k & 7.4k & 8.1k & 55 & 59 \\
fr$\leftrightarrow$it & 3.1M & 108M & 96.5M & 202k & 249k & 2k & 69k & 61k & 8.2k & 8.8k & 47 & 57 \\
fr$\leftrightarrow$ar & 5.0M & 152M & 152M & 290k & 320k  & 2k & 60k & 60k & 8.5k & 8.6k & 42 & 61 \\
fr$\leftrightarrow$es & 2.8M & 99.0M & 91.7M & 170k & 212k & 2k & 69k & 64k & 8.0k & 8.6k & 37 & 55 \\
fr$\leftrightarrow$de & 2.4M & 73.4M & 62.3M & 172k & 253k & 2k & 57k & 48k & 7.5k & 9.0k & 59 & 104 \\
fr$\rightarrow$zh & 3.0M & 98.5M & 76.3M & 199k & 168k & 2k & 67k & 51k & 8.0k & 5.9k & 51 & - \\
\hline
ja$\leftrightarrow$ko & 1.4M & 14.0M & 13.9M & 61.9k & 55.6k & 2k & 19k & 19k & 9.3k & 8.5k & 0 & 0 \\
\hline
nl$\rightarrow$fr & 3.0M & 74.8M & 84.7M & 446k & 260k & 2k & 49k & 55k & 7.9k & 7.5k & 150 & - \\
fa$\rightarrow$en &  795k & 21.7M & 20.2M & 166k & 147k & 2k & 54k  & 51k & 7.7k & 8.7k & 197 & -  \\
ja$\rightarrow$en & 1.3M & 28.0M & 22.0M & 24k & 87k & 2k & 41k & 32k & 6.2k & 7.3k & 3 & - \\
zh$\rightarrow$en  & 5.8M & 145M & 154M & 246k & 225k & 2k & 48k & 51k & 5.5k & 6.9k & 34 & - \\
\hline
\end{tabular}
\end{center}
\caption{\label{tab:corpora} Corpora statistics for each language pair (iso 639-1 2-letter code, expect for Portuguese Brazilian noted as "br"). All language pairs are bidirectional except nlfr, frzh, jaen, faen, enko, zhen. Columns 2-6 indicate the number of sentences, running words and vocabularies referred to training datasets while columns 7-11 indicate the number of sentences, running words and vocabularies referred to test datasets. Columns 12 and 13 indicate respectively the vocabulary of OOV of the source and target test sets. ({\it M} stand for milions, {\it k} for thousands). Since jako and enko are trained using BPE tokenization (see section \ref{ssec:tokenization}), there is no OOV.}
\end{table*}

Note that size of the corpus needs to be considered with the number of training periods since the neural network is continuously fed by sequences of sentence batches till the network is considered trained. In \newcite{WIPO}, authors mention using corpus of 5M sentences and training of 1.2M batches each having 40 sentences -- meaning basically that each sentence of the full corpus is presented 10 times to the training. In \newcite{GNMT}, authors mention 2M steps of 128 examples for English--French, for a corpus of 36M sentences, meaning about 7 iterations on the complete corpus. In our framework, for this release, we systematically extended the training up to 18 epochs and for some languages up to 22 epochs.

Selection of the optimal system is made after the complete training by calculating scores on independent test sets. As an outcome, we have seen different behaviours for different language pairs with similar training corpus size apparently connected to the language pair complexity. For instance, English--Korean training perplexity still decreases significantly between epoch 13 and 19 while Italian--English perplexity decreases marginally after epoch 10. For most languages, in our set-up, optimal systems are achieved around epoch 15.

We did also some experiment on the corpus size. Intuitively, since NMT systems do not have the memorizing capacity of PBMT engines, the fact that the training use 10 times 10M sentence corpus, or 20 times 5M corpus should not make a huge difference. In one of the experiment, we compared training on a 5M corpus trained over 20 epochs for English to/from French, and the same 5M corpus for only 10 epochs, followed by 10 additional epochs on additional 5M corpus. The 10M being completely homogeneous. In both directions, we observe that the $5\times 10+5\times 10$ training is completing with a score improvement of $0.8-1.2$ compared to the $5\times 20$ showing that the additional corpus is managing to bring a meaningful improvement. This observation leads to a more general question about how much corpus is needed to actually build a high quality NMT engine (learn the language), the role and timing of diversity in the training and whether the incremental gain could not be substituted by terminology feeding (learn the lexicon).

\section{Technology}
\label{sec:technology}

In this section we account for several experiments that improved different aspects of our translation engines. Experiments range from preprocessing techniques to extend the network with the ability to handle named entities, to use multiple word features and to enforce the attention module to be more like word alignments. We also report on different levels of translation customization.


\subsection{Tokenization}
\label{ssec:tokenization}

All corpora are preprocessed with an in-house toolkit.
We use standard token separators (spaces, tabs, etc.) as well as a set of language-dependent linguistic rules.
Several kinds of entities are recognized (url and number) replacing its content by the appropriate place-holder.
A postprocess is used to detokenize translation hypotheses, where the original raw text format is regenerated following equivalent techniques.

For each language, we have access to language specific tokenization and normalization rules.
However, our preliminary experiments showed that there was no obvious gain of using these language specific tokenization patterns, and that some of the hardcoded rules were actually degrading the performance.
This would need more investigation, but for the release of our first batch systems, we used a generic tokenization model for most of the languages except Arabic, Chinese and German.
In our past experiences with Arabic, separating segmentation of clitics was beneficial, and we retained the same procedure.
For German and Chinese, we used in-house compound splitter and word segmentation models, respectively.

In our current NMT approach, vocabulary size is an important factor that determines the efficiency and the quality of the translation system;
a larger vocabulary size correlates directly to greater computational cost during decoding, whereas low coverage of vocabulary leads to severe out-of-vocabulary (OOV) problems, hence lowering translation quality.

In most language pairs, our strategy combines a vocabulary shortlist and a placeholder mechanism, as described in Sections \ref{ssec:features} and \ref{ssec:namedentities}.
This approach, in general, is a practical and linguistically-robust option to addressing the fixed vocabulary issue, since we can take the full advantage of internal manually-crafted dictionaries and customisized user dictionaries (UDs).

A number of previous work such as character-level \cite{Chung2016}, hybrid word-character-based \cite{Luong2016} and subword-level \cite{Sennrich2016} address issues that arise with morphologically rich languages such as German, Korean and Chinese.
These approaches either build accurate open-vacabulary word representations on the source side or improve translation models' generative capacity on the target side.
Among those approaches, subword tokenization yields competitive results achieving excellent vocabulary coverage and good efficiency at the same time.

For two language pairs: enko and jaen, we used source and target sub-word tokenization (BPE, see \cite{Sennrich2016}) to reduce the vocabulary size but also to deal with rich morphology and spacing flexibility that can be observed in Korean.
Although this approach is very seducing by its simplicity and also used systematically in \cite{GNMT} and \cite{WIPO}, it does not have significant side effects (for instance generation of impossible words) and is not optimal to deal with actual word morphology - since the same suffix ({\it josa} in Korea) depending on the frequency of the word ending it is integrated with, will be splitted in multiple representations.
Also, in Korean, these {\it josa}, are is an ``agreement'' with the previous syllabus based on their final endings: however such simple information is not explicitely or implicitely reachable by the neural network.

The sub-word encoding algorithm \emph{Byte Pair Encoding} (BPE) described by \newcite{Sennrich2016} was re-implemented in C++ for further speed optimization.

%
%
%
%

\subsection{Word Features}
\label{ssec:features}

\newcite{DBLP:journals/corr/SennrichH16} showed that using additional input features improves translation quality. Similarly to this work, we introduced in the framework the support for an arbitrary number of discrete word features as additional inputs to the encoder. Because we do not constrain the number of values these features can take at the same time, we represent them with continuous and normalized vectors. For example, the representation of a feature $f$ at time-step $t$ is:

\begin{equation}
  x_i^{(t)} =
  \begin{cases}
    \frac{1}{n_f} & \quad \text{ if } f \text{ takes the } i^{th}\text{ value} \\
    0 & \quad \text{otherwise} \\
  \end{cases}
\end{equation}

where $n_f$ is the number of possible values the feature $f$ can take and with $x^{(t)} \in \mathbb{R}^{n_f}$.

These representations are then concatenated with the word embedding to form the new input to the encoder.

We extended this work by also supporting additional features on the target side which will be predicted by the decoder. We used the same input representation as on the encoder side but shifted the features sequence compared to the words sequence so that the prediction of the features at time-step $t$ depend on the word at time-step $t$ they annotate. Practically, we are generating feature at time $t+1$ for the word generated at time $t$.

To learn these target features, we added a linear layer to the decoder followed by the softmax function and used the mean square error criterion to learn the correct representation.

\begin{figure}
  \begin{tikzpicture}
    \begin{axis}[
      xlabel={Perplexity},
      ylabel={BLEU},
      xmin=4, xmax=17,
      ymin=14, ymax=29,
      xtick={6,8,10,12,14,16},
      ytick={16,18,20,22,24,26,28},
      width=0.45\textwidth,
      ymajorgrids=true,
      grid style=dashed,
      legend pos=north east,
      legend cell align=right,
      legend style={font=\tiny, draw=none},
      ]
      \addplot[
      color=black,dashed
      ]
      coordinates {
        (13.59, 15.21)
        (9.29, 19.28)
        (8.11, 23.13)
        (7.09, 24.59)
        (6.84, 24.31)
        (6.56, 25.06)
        (6.31, 25.13)
        (6.29, 25.37)
        (5.84, 26.26)
        (5.66, 26.45)
        (5.61, 26.97)
      };
      \addplot[
      color=blue,
      mark=diamond,
      ]
      coordinates {
        (14.31, 13.3)
        (9.57, 20.31)
        (7.98, 23.02)
        (7.33, 24.3)
        (7.05, 24.62)
        (6.68, 25.3)
        (6.46, 25.29)
        (6.26, 25.32)
        (6.25, 25.4)
        (5.77, 26.75)
        (5.59, 27.28)
        (5.53, 27.44)
        (5.51, 27.31)
      };
      \addplot[
      color=red,
      mark=square,
      ]
      coordinates {
        (14.84, 17.65)
        (10.21, 21.63)
        (8.86, 23.22)
        (8.1, 24.74)
        (6.93, 26.22)
        (6.59, 27.15)
        (6.46, 27.19)
        (6.41, 27.25)
        (6.37, 27.36)
        (6.37, 27.44)
        (6.36, 27.54)
        (6.36, 27.54)
      };
      \addplot[
      color=red,
      mark=square*,
      ]
      coordinates {
        (28.13, 10.44)
        (12.15, 21.07)
        (9.88, 22.39)
        (8.88, 24.52)
        (8.37, 25.44)
        (8.17, 25.95)
        (7.59, 26.16)
        (7.59, 26.16)
        (6.93, 27.29)
        (6.72, 27.99)
        (6.62, 28.2)
        (6.59, 28.09)
      };
      \legend{baseline, source case, src\&tgt, src\&tgt+embed}
    \end{axis}
  \end{tikzpicture}
  \caption{Comparison of training progress (perplexity/BLEU) with/without source (src) and target (tgt) case features, with/without feature embedding (embed) on WMT2013 test corpus for English-French. Score is calculated on lowercase output. The perplexity increases when the target features are introduced because of the additional classification problem. We also notice a noticeable increase in the score when introducing the features, in particular the target features. So these features do not simply help to reduce the vocabulary, but also by themselves help to structure the NN decoding model.}
  \label{fig:feature_training}
\end{figure}
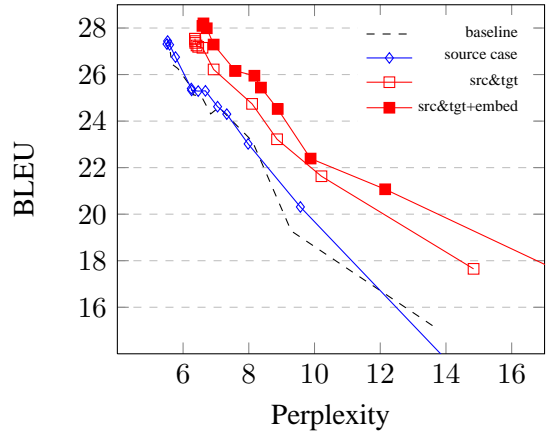

For this release, we only used case information as additional feature. It allows us to work with a lowercased vocabulary and treat the recasing as a separate problem. We observed that the use of this simple case feature in source and target does improve the translation quality as illustrated in Figure \ref{fig:feature_training}. Also, we compared the accuracy of the induced recasing with other recasing frameworks (SRI disamb, and in-house recasing tool based on $n$-gram language models) and observed that the prediction of case by the NN was higher than using external recaser, which was expected since NN has access to source in addition to the source sentence context, and target sentence history.

\subsection{Named Entities (NE)}
\label{ssec:namedentities}

SYSTRAN's RBMT and SMT translation engines utilize number and named entity (NE) module to recognize, protect, and translate NE entities.
Similarly, we used the same internal NE module to recognize numbers and named entities in the source sentence and
temporarily replaced them with their corresponding placeholders (Table \ref{tab:neplaceholders}).

\begin{table}
\small
\begin{center}
\begin{tabular}{|c|c|}
\hline  NE type & Placeholder \\
\hline
Number & \_\_ent\_numeric \\
Measurement & \_\_ent\_numex\_measurement  \\
Money & \_\_ent\_numex\_money  \\
\hline
Person (any) & \_\_ent\_person  \\
Title & \_\_ent\_person\_title  \\
First name & \_\_ent\_person\_firstname  \\
Initial & \_\_ent\_person\_initials  \\
Last name & \_\_ent\_person\_lastname  \\
Middle name  & \_\_ent\_person\_middlename \\
 \hline
Location & \_\_ent\_location  \\
Organization & \_\_ent\_organization  \\
Product & \_\_ent\_product \\
Suffix & \_\_ent\_suffix  \\
 \hline
Time expressions & \_\_ent\_timex\_expression  \\
Date  & \_\_ent\_date  \\
Date (day) & \_\_ent\_date\_day  \\
Date (Month) & \_\_ent\_date\_month  \\
Date (Year) & \_\_ent\_date\_year  \\
Hour & \_\_ent\_hour \\
\hline
\end{tabular}
\end{center}
\caption{\label{tab:neplaceholders} Named entity placeholder types}
\end{table}

Both the source and the target side of the training dataset need to be processed for NE placeholders.
To ensure the correct entity recognition, we cross-validate the recognized entity across parallel dataset, that is:
a valid entity recognition in a sentence should have the same type of entity in its parallel pair and the word or phrase covered by the entities need to be aligned to each other.
We used \texttt{fast\_align} \cite{Dyer2013fastAlign} to automatically align source words to target words.

In our datasets, generally about one-fifth of the training instances contained one or more NE placeholders.
Our training dataset consists of sentences with NE placeholders as well as sentences without them
to be able to handle both instantiated and recognized entity types.

Per source sentence, a list of all entities, along with their translations in the target language, if available, are returned by our internal NE recognition module.
The entities in the source sentence is then replaced with their corresponding NE placeholders.
During beam search, we make sure that an entity placehoder is translated by itself in the target sentence.
When the entire target sentence is produced along with the attention weights that provide soft alignments back to the original source tokens,
Placeholders in the target sentences are replaced with either the original source string or its translation.

The substitution of the NE placeholders with their correct values needs language pair-specific considerations.
In Figure \ref{tab:structuralNEexample}, we show that even the handling of Arabic numbers cannot be straight-forward as copying the original value in the source text.

\begin{table}
\small
\begin{center}
\begin{tabular}{c|cc}
& \multicolumn{2}{c}{En $\rightarrow$ Ko} \\
\hline
& \multicolumn{2}{c}{Train} \\
\hline
train data & 25 billion & 250억 \\
entity-replaced &  \_\_ent\_numeric billion & \_\_ent\_numeric 억 \\
\hline
& \multicolumn{2}{c}{Decode} \\
\hline
input data & 1.4 billion &  - \\
entity-replaced & \_\_ent\_numeric billion &  - \\
translated & - & \_\_ent\_numeric 억 \\
naive substition & - & 1.4억 \\
expected result & -  &  14억 \\
\end{tabular}
\end{center}
\caption{\label{tab:structuralNEexample} Examples of English and Korean number expressions where naive recognition and substitution fails. Even if the model correctly produces correct placeholders, simply copying the original value will result in incorrect translation. These kind of \emph{structural entities} need language pair-specific treatment.}
\end{table}

\subsection{Guided Alignments}
\label{ssec:guidedalignments}

We re-reimplemented \emph{Guided alignment} strategy described in \newcite{DBLP:journals/corr/ChenMKP16}.
\emph{Guided alignment} enforces the attention weights to be more like alignments in the traditional sense (e.g. IBM4 viterbi alignment) where the word alignments explicitly indicate that source words aligned to a target word are translation equivalents.

Similarly to the previous work, we created an additional criterion on attention weights, $L_{ga}$, such that
the difference in the attention weights and the reference alignments is treated as an error and directly and additionally optimize the output of the attention module.
\begin{equation}
L_{ga}(A,\alpha) = \dfrac{1}{T} \cdot \sum_{t}{\sum_{s}{(A_{st} - \alpha_{st})^{2}}} \nonumber
\end{equation}
The final loss function for the optimization is then:
\begin{equation}
L_{total} = w_{ga} \cdot L_{ga}(A,\alpha) + (1-w_{ga}) \cdot L_{dec}(y,x) \nonumber
\end{equation}
where $A$ is the alignment matrix, $\alpha$ attention weights, $s$ and $t$ indicating the indices in the source and target sentences,
and $w_{ga}$ the linear combination weight for the guided alignment loss.

\newcite{DBLP:journals/corr/ChenMKP16} also report that decaying $w_{ga}$, thereby gradually reducing the influence from guided alignment, over the course of training found to be helpful on certain datasets. When guided alignment decay is enabled, $w_{ga}$ is gradually reduced at this rate after each epoch from the beginning of the training.

Without searching for the optimal parameter values, we simply took following configurations from the literature:
mean square error (MSE) as loss function, 0.5 as the linear-combination weight for guided alignment loss and the cross-entropy loss for decoder, and 0.9 for decaying factor for guided alignment loss.

For the alignment, we again utilized \texttt{fast\_align} tool.
We stored alignments in sparse format\footnote{Compressed Column Storage (CCS)} to save memory usage, and for each minibatch a dense matrix is created for faster computation.

Applying such a constraint on attention weights can help locate the original source word more accurately, which we hope to benefits
better NE placeholder substitutions and especially unknown word handling.

\begin{figure}
  \begin{tikzpicture}
    \begin{axis}[
      xlabel={Epoch},
      ylabel={BLEU},
      xmin=1, xmax=19,
      ymin=46, ymax=52.7,
      xtick={1,3,5,7,9,11,13,15,17,19},
      ytick={46,47,48,49,50,51,52,53},
      width=0.45\textwidth,
      ymajorgrids=true,
      grid style=dashed,
      legend pos=south east,
      legend cell align=left,
      legend style={font=\tiny, draw=none},
      ]
      \addplot[
      color=black,
      mark=triangle,
      ]
      coordinates {
        (3, 47.36)
        (4, 48.52)
        (5, 48.96)
        (6, 48.84)
        (7, 49.54)
        (8, 49.02)
        (9, 49.58)
        (10, 50.46)
        (11, 51.37)
        (13, 51.05)
        (14, 51.49)
        (15, 51.52)
        (16, 51.69)
        (17, 51.88)
        (18, 51.86)
      };
      \addplot[
      color=blue,
      mark=diamond,
      ]
      coordinates {
        (2, 47.64)
        (3, 49.32)
        (4, 49.80)
        (5, 50.45)
        (6, 50.80)
        (7, 51.05)
        (8, 50.75)
        (9, 50.81)
        (10, 51.30)
        (11, 51.54)
        (12, 52.08)
        (13, 52.63)
      };
      \addplot[
      color=red,
      mark=square,
      ]
      coordinates {
        (2, 47.57)
        (3, 49.25)
        (4, 50.07)
        (5, 50.85)
        (6, 51.40)
        (7, 51.83)
        (8, 52.03)
      };
      \legend{Baseline, Guided alignment, Guided alignment with decay}
    \end{axis}
  \end{tikzpicture}
  \caption{Effect of guided alignment. This particular learning curve is from training an attention-based model with 4 layers of bidirectional LSTMs with 800 dimension on a 5 million French to English dataset.}
  \label{fig:guidedalignment_bleu}
\end{figure}
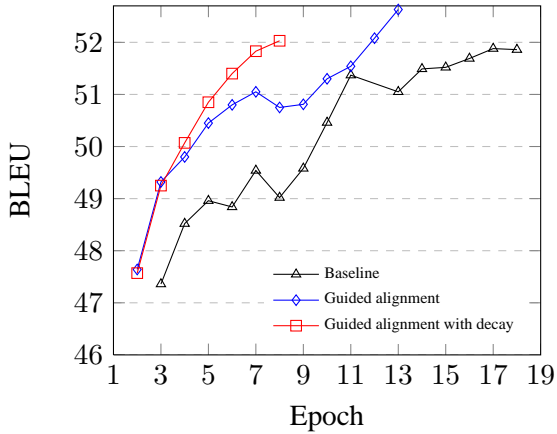

In Figure \ref{fig:guidedalignment_bleu}, we see the effects of guided alignment and its decay rate on our English-to-French generic model.
Unfortunately, this full comparative run was disrupted by a power outage and we did not have time to relaunch, however we can still clearly observe that up to the initial 5 epochs, guided alignment, with or without decay, provides rather big boosts over the baseline training. After 5 epochs, the training with decay slow down compare to the training without, which is rather intuitive: the guided alignment is indeed in conflict with the attention learning. What would remain to be seen, is if at the training the training, the baseline and the guided alignment with decay are converging.

\subsection{Politeness Mode}
\label{ssec:politeness}

Many languages have ways to express politeness and deference towards people being reffered to in sentences.
In Indo-European languages, there are two pronouns corresponding to the English You;
it is called the \textit{T-V} distinction between the informal Latin pronoun \textit{tu} (\textbf{T}) and the polite Latin pronoun \textit{Vos} (\textbf{V}).
Asian languages, such as Japanese and Korean, make an extensive use of honorifics (respectful words), words that are usually appended to the ends of names or pronouns to indicate the relative ages and social positions of the speakers.
Expressing politeness can also impact the vocabulary of verbs, adjectives, and nouns used, as well as sentence structures.

Following the work of \newcite{sennrich-haddow-birch_2016_NAACLHLT}, we implemented a politeness feature in our NMT engine: a special token is added to each source sentence during training, where the token indicates the politeness mode observed in the target sentence.
Having such an ability to specify the politeness mode is very useful especially when translating from a language where politeness is not expressed, e.g. English, into where such expressions are abundant, e.g. Korean, because it provides a way of customizing politeness mode of the translation output.\\
Table \ref{tab:politenessenkoexample} presents our English-to-Korean NMT model trained with politeness mode, and it is clear that the proper verb endings are generated according to the user selection. After a preliminary evaluation on a small testset from English to Korean, we observed 70 to 80\% accuracy of the politeness generation (Table \ref{tab:politenesseval}).
We also noticed that 86\% of sentences (43 out of 50) have exactly the same meaning preserved across different politeness modes.

This simple approach, however, comes at a small price, where sometimes the unknown replacement scheme tries to copy the special token in the target generation. A more appropriate approach that we plan to switch to in our future trainings is to directly feed the politeness mode into the sentential representation of the decoder.

\begin{table*}
\small
\begin{center}
\begin{tabular}{l}
\hline
\bf{En}:\\
\hspace{5mm} A senior U.S. treasury official is urging china to move faster on making its currency more flexible. \\
\hline
\textbf{Ko} with \textbf{Formal} mode: \\
\hspace{5mm} 미국 재무부 고위 관계자는 중국이 위안화의 융통성을 더 유연하게 만들기 위해 더 빨리 움직일 것을 {\color{red}{촉구했습니다.}} \\
\hline
\textbf{Ko} with \textbf{Informal} mode: \\
\hspace{5mm} 미국 재무부 고위 관계자는 중국이 위안화를 좀더 유연하게 만들기 위해 더 빨리 움직일 것을 {\color{red}{촉구하고 있어요.}} \\
\hline
\end{tabular}
\end{center}
\caption{\label{tab:politenessenkoexample} A translation examples from an En-Ko system where the choice of different politeness modes affects the output.}
\end{table*}

\begin{table}
\begin{center}
\begin{tabular}{ c r r r }
\hline
Mode 	& Correct & Incorrect		& Accuracy \\
\hline
Formal	& 30	& 14 	& 68.2\% \\
Informal	& 35	& 9	& 79.5\% \\
\hline
\end{tabular}
\end{center}
\caption{\label{tab:politenesseval} Accuracy of generating correct \emph{Politeness mode} of an English-to-Korean NMT system. The evaluation was carried out on a set of 50 sentences only; 6 sentences were excluded from evaluation because neither the original nor their translations contained any verbs.}
\end{table}

\subsection{Customization}
\label{ssec:adaptation}

Domain adaptation is a key feature for our customers---it generally encompasses terminology, domain and style adaptation, but can also be seen as an extension of translation memory for human post-editing workflows.

SYSTRAN engines integrate multiple techniques for domain adaptation, training full new in-domain engines, automatically post-editing an existing translation model using translation memories, extracting and re-using terminology.
With Neural Machine Translation, a new notion of ``specialization'' comes close to the concept of incremental translation as developed for statistical machine translation like \cite{ortiz2010online}.

\subsubsection{Generic Specialization}
\label{sssec:specialization}

Domain adaptation techniques have successfully been used in Statistical Machine Translation. It is well known that a system optimized on a specific text genre obtains higher accuracy results than a ``generic'' system. The adaptation process can be done before, during or after the training process. Our preliminary experiments follow the latter approach. We incrementally adapt a Neural MT ``generic'' system to a specific domain by running additional training epochs over newly available in-domain data.

Adaptation proceeds incrementally when new in-domain data becomes available, generated by human translators while post-editing, which is similar to the Computer Aided Translation framework described in \cite{Cettolo2014}.

We experiment on an English-to-French translation task. The generic model is a subsample of the corpora made available for the WMT15 translation task \cite{wmt15}. Source and target NMT vocabularies are the 60k most frequent words of source and target training datasets. The in-domain data is extracted from the European Medical Agency (EMEA) corpus. Table \ref{tab:XP_DA} shows some statistics of the corpora used in this experiment.

\begin{table}[h!]
\centering
\scriptsize
 \begin{tabular}{llccc}
\hline 
 \textit{Type} & \textit{Corpus}& $\#$ \textit{lines}  &  $\#$ \textit{src tok (EN)} & $\#$ \textit{tgt tok (FR)} \\
\hline 
Train & Generic & 1M     & 24M & 26M  \\
        & EMEA   & 4,393 & 48k & 56k \\
\hline 
Test &  EMEA & 2,787 & 23k & 27k    \\
\hline 
 \end{tabular}
 \caption{\label{tab:XP_DA}Data used to train and adapt the generic model to a specific domain. The test corpus also belongs to the specific domain.}
\end{table}

Our preliminary results show that incremental adaptation is effective for even limited amounts of in-domain data (nearly 50k additional words). Constrained to use the original ``generic'' vocabulary, adaptation of the models can be run in a few seconds, showing clear quality improvements on in-domain test sets. 

\begin{figure}[h!]
  \begin{tikzpicture}
    \begin{axis}[
      xlabel={Additional epochs},
      ylabel={BLEU},
      xmin=0, xmax=5,
      ymin=28, ymax=37,
      xtick={0,1,2,3,4,5},
      ytick={29,30,31,32,33,34,35,36},
      width=0.45\textwidth,
      ymajorgrids=true,
      grid style=dashed,
      legend pos=south east,
      legend cell align=left,
      legend style={font=\small, draw=none},
      ]
	
      \addplot[
      color=red,
      mark=square,
      ]
      coordinates {
        (0, 29.01)
        (1, 34.00)
        (2, 33.78)
        (3, 34.03)
        (4, 33.92)
        (5, 33.92)
      };
     
      \addplot[
      color=blue,
      ]
      coordinates {
        (-1, 34.91)
        (6, 34.91)
      };	
	\legend{adapt, full}
    \end{axis}
  \end{tikzpicture}
  \caption{Adaptation with In-Domain data.}
  \label{fig:domainAdaptationFewData}
\end{figure}
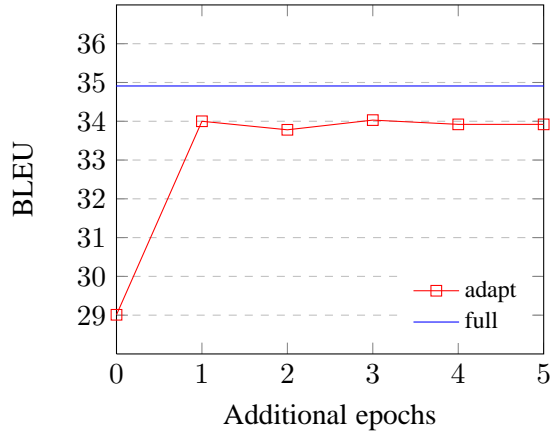

Figure \ref{fig:domainAdaptationFewData} compares the accuracy (BLEU) of two systems: full is trained after concatenation of generic and in-domain data; adapt is initially trained over generic data (showing a BLEU score of $29.01$ at epoch $0$) and adapted after running several training epochs over {\it only} the in-domain training data. Both systems share the same "generic" NMT vocabularies. As it can be seen the adapt system improves drastically its accuracy after a single additional training epoch, obtaining a similar BLEU score than the full system (separated by $.91$ BLEU). Note also that each additional epoch using the in-domain training data takes less than $50$ seconds to be processed, while training the full system needs more than $17$ hours. 

Results validate the utility of the adaptation approach. A human post-editor would take advantage of using new training data as soon as it becomes available, without needing to wait for a long full training process. However, the comparison is not entirely fair since full training would allow to include the in-domain vocabulary in the new full model, what surely would result in an additional accuracy improvement.

\subsubsection{Post-editing Engine}
\label{sssec:postedition}

Recent success of Pure Neural Machine Translation has led to the application of this technology to various related tasks and in particular to the Automatic Post-Editing (APE). The goal of this task is to simulate the behavior of a human post-editor, correcting translation errors made by a MT system.

Until recently, most of the APE approaches have been based on phrase-based SMT systems, either monolingual (MT target to human post-edition) \cite{Simard07statisticalphrase-based} or source-aware \cite{bechara2011statistical}. For many years now SYSTRAN has been offering a hybrid Statistical Post-Editing (SPE) solution to enhance the translation provided by its rule-based MT system (RBMT) \cite{dugast-senellart-koehn:2007:WMT}.

Following the success of Neural Post-Editing (NPE) in the APE task of WMT’16 \cite{DBLP:journals/corr/Junczys-Dowmunt16}, we have run a series of experiments applying the neural approach in order to improve the RBMT system output. As a first experiment, we compared the performance of our English-to-Korean SPE system trained on technical (IT) domain data to two NPE systems trained on the same data: monolingual NPE and multi-source NPE, where the input language and the MT hypothesis sequences have been concatenated together into one input sequence (separated by a special token).


\begin{figure}
  \begin{tikzpicture}
    \begin{axis}[
      xlabel={Epoch},
      ylabel={BLEU},
      xmin=1, xmax=13,
      ymin=0, ymax=54,
      xtick={1,3,5,7,9,11,13},
      ytick={0,10,20,30,40,50},
      width=0.45\textwidth,
      ymajorgrids=true,
      grid style=dashed,
      legend pos=south east,
      legend cell align=left,
      legend style={font=\tiny, draw=none},
      ]

      \addplot[
      color=blue,
      mark=diamond,
      ]
      coordinates {
        (1, 3.13)
        (2, 12.48)
        (3, 23.47)
        (4, 35.53)
        (5, 39.78)
        (6, 41.98)
        (7, 44.01)
        (8, 43.72)
        (9, 48.88)
        (10, 51.09)
        (11, 52.17)
        (12, 52.32)
        (13, 52.79)
      };

      \addplot[
      color=red,
      mark=square,
      ]
      coordinates {
        (1, 6.87)
        (2, 27.85)
        (3, 34.97)
        (4, 38.84)
        (5, 40.69)
        (6, 43.12)
        (7, 42.67)
        (8, 45.20)
        (9, 49.13)
        (10, 50.92)
        (11, 51.22)
        (12, 52.04)
        (13, 52.38)
      };

      \addplot[
      color=black,
      mark=triangle,
      ]
      coordinates {
        (1, 3.14)
        (2, 11.29)
        (3, 21.05)
        (4, 26.56)
        (6, 29.60)
        (8, 32.19)
        (9, 34.79)
        (10, 37.03)
        (11, 37.77)
        (12, 37.49)
        (13, 37.25)
      };

      \addplot[
      color=green,
      dashed,
      ]
      coordinates {
        (0, 33.55)
        (14, 33.55)
      };

      \addplot[
      color=brown,
      dotted,
      ]
      coordinates {
        (0, 8.11)
        (14, 8.11)
      };

      \legend{NMT, NPE multi-source, NPE, SPE, RBMT}
    \end{axis}
  \end{tikzpicture}
  \caption{Accuracy results of RBMT, NMT, NPE and NPE multi-source.}
  \label{fig:zophrnn_npe}
\end{figure}
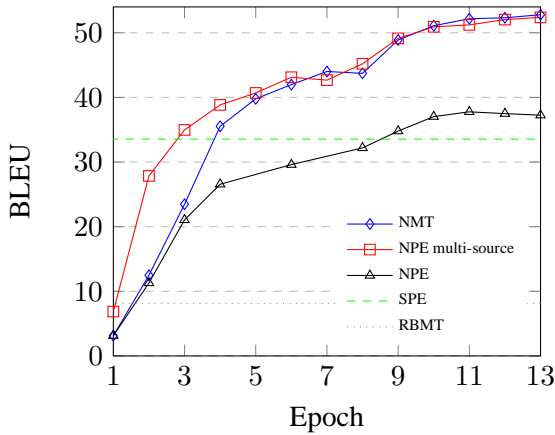

Figure \ref{fig:zophrnn_npe} illustrates the accuracy (BLEU) results of four different systems at different training epochs. The RBMT system performs poorly, confirming the importance of post-editing. Both NPE systems clearly outperform SPE. It can also be observed that adding source information even in a simplest way possible (NPE multi-source), without any source-target alignment, considerably improves NPE translation results.

The system performing NPE multi-source obtains similar accuracy results than pure NMT. What can be seen is that NPE multi-source essentially employs the information from the original sentence to produce translations. However, notice that benefits of utilizing multiple inputs from different sources is clear at earlier epochs while once the model parameters converge, the difference in performances of NMT and NPE multi-source models become negligible.

Further experiments are currently being conducted aiming at finding more sophisticated ways of combining the original source and the MT translation in the context of NPE.

\section{Performance}
\label{sec:performance}

As previously outlined, one of the major drawbacks of NMT engines is the need for cutting-edge hardware technology to face the enormous computational requirements at training and runtime.

Regarding training, there are two major issues: the full training time and the required computation power, i.e. the server investment. For this release, most of our trainings have been running on single GTX GeForce 1080 GPU (about \$2.5k) while in \cite{GNMT}, authors mention using 96 K80 GPU for a full week for training one single language pair (about \$250k). On our hardware, full training on 2x5M sentences (see section \ref{sec:corpora}) took a bit less than one month. 

A reasonnable target is to maintain training time for any language pair under one week and keeping reasonable investment so that the full research community can have competitive trainings but also indirectly so that all of our customers can benefit from the training technology. 
To do so, we need to better leverage multiple GPUs on a single server which is on-going engineering work. We also need to continue on exploring how to learn more with less data. For instance, we are convinced that injecting terminology as part of the training data should be competitive with continuing adding full sentences. 

Also, shortening training cycle can also be achieved by better control of the training cycle.
We have shown that multiple features are boosting the training pace, and if going to bigger network is clearly improving performance. For instance, we are using a bidirectional 4 layer RNN in addition to our regular RNN, but in \newcite{GNMT}, authors mention using bidirectional RNN only for the first layer. We need to understand more these phenomena and restrict to the minimum to reduce the model size. 

Finally, work on specialization described in sections \ref{sssec:specialization} and \ref{ssec:distillation} are promising for long term maintenance: we could reach a point where we do not need to retrain from scratch but continuously improve existing model and use teacher models to boost initial trainings.

Regarding runtime performance, we have been exploring the following areas and are reaching today throughputs compatible with production requirement not only using GPUs but also using CPU and we report our different strategies in the following sub-sections.

\subsection{Pruning}
\label{ssec:pruning}

Pruning the parameters of a neural network is a common technique to reduce its memory footprint. This approach has been proven efficient for the NMT tasks in \newcite{see2016compression}. Inspired by this work, we introduced similar pruning techniques in \texttt{seq2seq-attn}. We reproduced that models parameters can be pruned up to 60\% without any performance loss after retraining as shown in Table \ref{tab:pruningbleu}.

\begin{table}
  \begin{center}
    \begin{tabular}{ | l | l |}
      \hline
      Model & BLEU \\ \hline
      baseline & 49.24 \\
      40\% pruned & 48.56 \\
      50\% pruned & 47.96 \\
      60\% pruned & 45.90 \\
      70\% pruned & 38.38 \\
      \hline
      60\% pruned and retrained & \textbf{49.26} \\ \hline
    \end{tabular}
    \caption{\label{tab:pruningbleu} BLEU scores of pruned models on an internal test set.}
  \end{center}
\end{table}

\begin{figure}
  \begin{center}
    \includegraphics[width=0.45\textwidth]{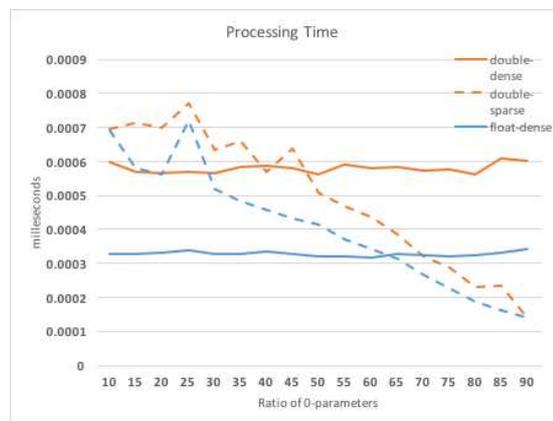}
    \caption{\label{fig:pruningPerf} Processing time to perform a $1000 \times 1000$ matrix multiplication on a single thread.}
  \end{center}
\end{figure}

\begin{figure}
  \begin{center}
    \includegraphics[width=0.45\textwidth]{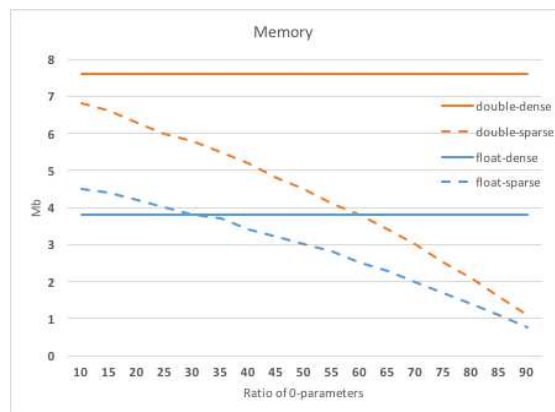}
    \caption{\label{fig:pruningMem} Memory needed to perform a $1000 \times 1000$ matrix multiplication.}
  \end{center}
\end{figure}

With a large pruning factor, neural network's weights can also be represented with sparse matrices. This implementation can lead to lower computation time but more importantly to a smaller memory footprint that allows us to target more environment. Figures \ref{fig:pruningPerf} and \ref{fig:pruningMem} show experiments involving sparse matrices using \emph{Eigen}\footnote{\url{http://eigen.tuxfamily.org}}. For example, when using the \texttt{float} precision, a multiplication with a sparse matrix already begins to take less memory when 35\% of its parameters are pruned.

Related to this work, we present in Section \ref{ssec:cdecoder} our alternative \emph{Eigen}-based decoder that allows us to support sparse matrices.

\subsection{Distillation}
\label{ssec:distillation}

Despite that surprisingly accurate, NMT systems need for deep networks in order to perform well. Typically, a 4-layer LSTM with 1000 hidden units per layer (4 x 1000) are used to obtain state-of-the-art results. Such models require cutting-edge hardware for training in reasonable time while inference becomes also challenging on standard setups, or on small devices such as mobile phones. Though, compressing deep models into smaller networks has been an active area of research. 

Following the work in ~\cite{knowledgedistillation2016}, we experimented sequence-level knowledge distillation in the context of an English-to-French NMT task. Knowledge distillation relies on training a smaller student network to perform better by learning the relevant parts of a larger teacher network. Hence, 'wasting'€™ parameters on trying to model the entire space of translations. Sequence-level is the knowledge distillation variant where the student model mimics the teacher's actions at the sequence-level. 

The experiment is summarized in 3 steps:
\begin{itemize}
\item train a teacher model on a source/reference training set,
\item use the teacher model to produce translations for the source training set,
\item train a student model on the new source/translation training set.
\end{itemize}

For our initial experiments, we produced {\it 35}-best translations for each of the sentences of the source training set, and used a normalized $n$-gram matching score computed at the sentence level, to select the closest translation to each reference sentence. The original training source sentences and their translated hypotheses where used as training data to learn a 2 x 300 LSTM network.

Results showed slightly higher accuracy results for a {\it 70\%} reduction of the number of parameters and a {\it 30\%} increase on decoding speed. In a second experiment, we learned a student model with the same structure than the teacher model. Surprisingly, the student clearly outperformed the teacher model by nearly $1.5$ BLEU. 

We hypothesize that the translation performed over the target side of the training set produces a sort of language normalization which is by construction very heterogeneous. Such normalization eases the translation task, being learned by not so deep networks with similar accuracy levels.


\subsection{Batch Translation}
\label{ssec:batch}

To increase translation speed of large texts, we support batch translation that works in addition to the beam search. It means that for a beam of size $K$ and a batch of size $B$, we forward $K \times B$ sequences into the model. Then, the decoder output is split across each batch and the beam path for each sentence is updated sequentially.

As sentences within a batch can have large variations in size, extra care is needed to mask accordingly the output of the encoder and the attention softmax over the source sequences.

\begin{figure}
  \begin{tikzpicture}
    \begin{axis}[
      xlabel={batch size},
      ylabel={tokens/s},
      xmin=0, xmax=100,
      ymin=0, ymax=500,
      xtick={0,20,40,60,80,100},
      ytick={0,100,200,300,400,500},
      width=0.45\textwidth,
      ymajorgrids=true,
      grid style=dashed,
      ]

      \addplot[
      color=blue,
      mark=square,
      ]
      coordinates {
        (1, 49.306026061)
        (2, 70.6692634376)
        (4, 103.987452199)
        (8, 169.0248843294)
        (16, 244.6862577047)
        (24, 304.0661522811)
        (32, 332.7213309962)
        (50, 373.0638327405)
        (100, 443.4885614854)
      };
    \end{axis}
  \end{tikzpicture}
  \caption{Tokens throughput when decoding from a student model (see section \ref{ssec:distillation}) with a beam of size 2 and using \texttt{float} precision. The experiment was run using a standard \emph{Torch} + \emph{OpenBLAS} install and 4 threads on a desktop \emph{Intel i7} CPU.}
  \label{fig:torch_batchdecoding}
\end{figure}
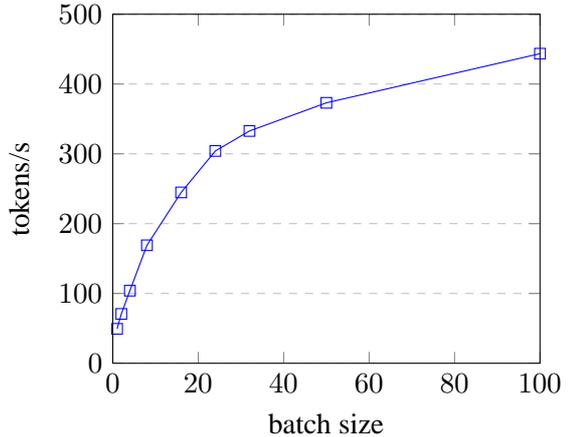

Figure \ref{fig:torch_batchdecoding} shows the speedup obtained using batch decoding in a typical setup.

\subsection{C++ Decoder}
\label{ssec:cdecoder}

While \emph{Torch} is a powerful and easy to use framework, we chose to develop an alternative C++ implementation for the decoding on CPU. It increases our control over the decoding process and open the path to further memory and speed improvements while making deployment easier.

Our implementation is graph-based and use \emph{Eigen} for efficient matrix computation. It can load and infer from \emph{Torch} models.

\begin{figure}
  \begin{tikzpicture}
    \begin{axis}[
      xlabel={threads},
      ylabel={tokens/s},
      xmin=1, xmax=8,
      ymin=0, ymax=100,
      xtick={1,2,4,8},
      ytick={0,20,40,60,80,100},
      legend pos=south east,
      width=0.45\textwidth,
      ymajorgrids=true,
      grid style=dashed,
      legend style={font=\small, draw=none},
      ]

      \addplot[
      color=red,
      mark=square,
      ]
      coordinates {
        (1, 44.4650817236)
        (2, 73.6574589715)
        (4, 90.5819721443)
        (8, 91.4920775657)
      };
      \addlegendentry{Eigen-based}

      \addplot[
      color=blue,
      mark=square,
      ]
      coordinates {
        (1, 49.3600570088)
        (2, 48.25180069)
        (4, 49.9501791081)
        (8, 45.4390747722)
      };
      \addlegendentry{Torch-based}
    \end{axis}
  \end{tikzpicture}
  \caption{Tokens throughput with a batch of size 1 in the same condition as Figure \ref{fig:torch_batchdecoding}.}
  \label{fig:eigen_torch_comparison}
\end{figure}
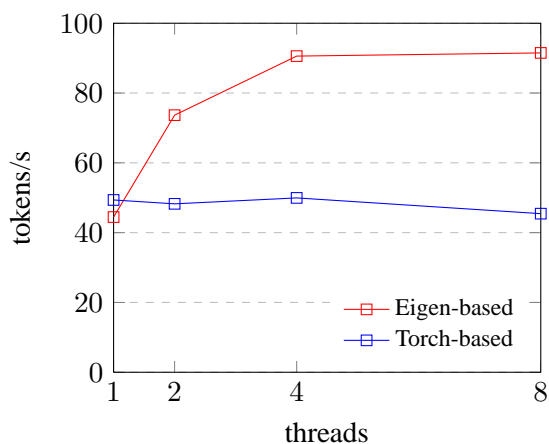

For this release, experiments show that the decoding speed is on par or faster than the \emph{Torch}-based implementation especially in a multi-threaded context. Figure \ref{fig:eigen_torch_comparison} shows the better use of parallelization of the \emph{Eigen}-based implementation.

\section{Evaluation}
\label{sec:evaluation}

Evaluation of machine translation has always been a challenge and subject to many papers and dedicated workshops \cite{Bojar2016}. While automatic metrics are now used as standard in the research world and have shown good correlation with human evaluation, ad-hoc human evaluation or productivity analysis metrics are rather used in the industry \cite{blain2011qualitative}.

As a translation solution company, even if automatic metrics are used through all the training process (and we give scores in the section \ref{ssec:eval_auto}), we care about human evaluation of the results. \newcite{GNMT} mention human evaluation but simultaneously cast a doubt on the referenced human to translate or evaluate. In this context, the claim ``{\it almost indistinguishable with human translation}'' is at the same time strong but also very vague. On our side, we have observed during all our experiments and preparation of specialized models, unprecedented level of quality, and contexts where we could claim ``super human'' translation quality.

However, we need to be very carefully defining the tasks, the human that are being compared to, and the nature of the evaluation. For evaluating technical translation, the nature of the evaluation is somewhat easy and really depending on the user expectation: is the meaning properly conveyed, or is the sentence faster to post-edit than to translate. Also, to avoid doubts about integrity or competency of the evaluation we sub-contracted the task to CrossLang, a company specialized in machine translation evaluation. The test protocol was defined collaboratively and for this first release, we decided to perform ranking of different systems, and we present in the section \ref{ssec:eval_human} the results obtained on two very different language pairs: English to/from French, and English to Korean. 

Finally, in the section \ref{ssec:eval_qualitative}, we also present some qualitative evaluation results showing specificities of the Neural Machine Translation.

\subsection{Automatic Scoring and system comparison}
\label{ssec:eval_auto}

\begin{figure*}
  \begin{tikzpicture}
    \begin{axis}[
      xlabel={Perplexity},
      ylabel={BLEU},
      xmin=2.3, xmax=6,
      ymin=25, ymax=56,
      xtick={2,3,4,5,6,7,8,9},
      ytick={20,25,30,35,40,45,50,55},
      width=1.0\textwidth,
      ymajorgrids=true,
      xmajorgrids=true,
      grid style=dashed,
nodes near coords align={right},
every node near coord/.append style={font=\small, inner sep=1pt},
      ]

\addplot[
   only marks,
    black,
    mark=.,
    nodes near coords, 
    point meta=explicit symbolic, 
    ] table [
     meta index=2 
     ] {
PPL       BLEU       label
2.79 43.65 enar
2.72 46.70 aren
3.63 33.72 ende
3.33 41.25 deen
3.08 40.55 ennl
3.25 42.42 nlen
3.08 42.82 enit
3.28 43.52 iten
2.68 46.20 enbr
2.67 48.91 bren
2.75 45.24 enes
2.94 45.58 esen
3.50 37.87 frar
3.28 36.47 arfr
3.06 43.55 frit
2.55 48.56 itfr
3.82 31.45 frde
2.99 39.97 defr
3.70 38.88 frbr
2.96 44.23 brfr
2.84 47.15 fres
2.49 49.58 esfr
5.72 31.67 frzh
2.57 45.79 zhen
2.40 51.04 nlfr
2.46 54.88 enfr
2.91 53.21 fren
4.85 33.41 jako
5.62 27.30 koja
3.84 34.19 faen
};

    \end{axis}
  \end{tikzpicture}
  \caption{Perplexity and BLEU scores obtained by NMT engines for all language pairs. Perplexity and BLEU scores were calculated respectively on validation and test sets. Note that English-to-Korean and Japanese-to-English systems are not shown in this plot achieving respectively ($18.94$, $16.50$) and ($8.82$, $19.99$) scores for perplexity and BLEU.}
  \label{fig:bleuppl}
\end{figure*}
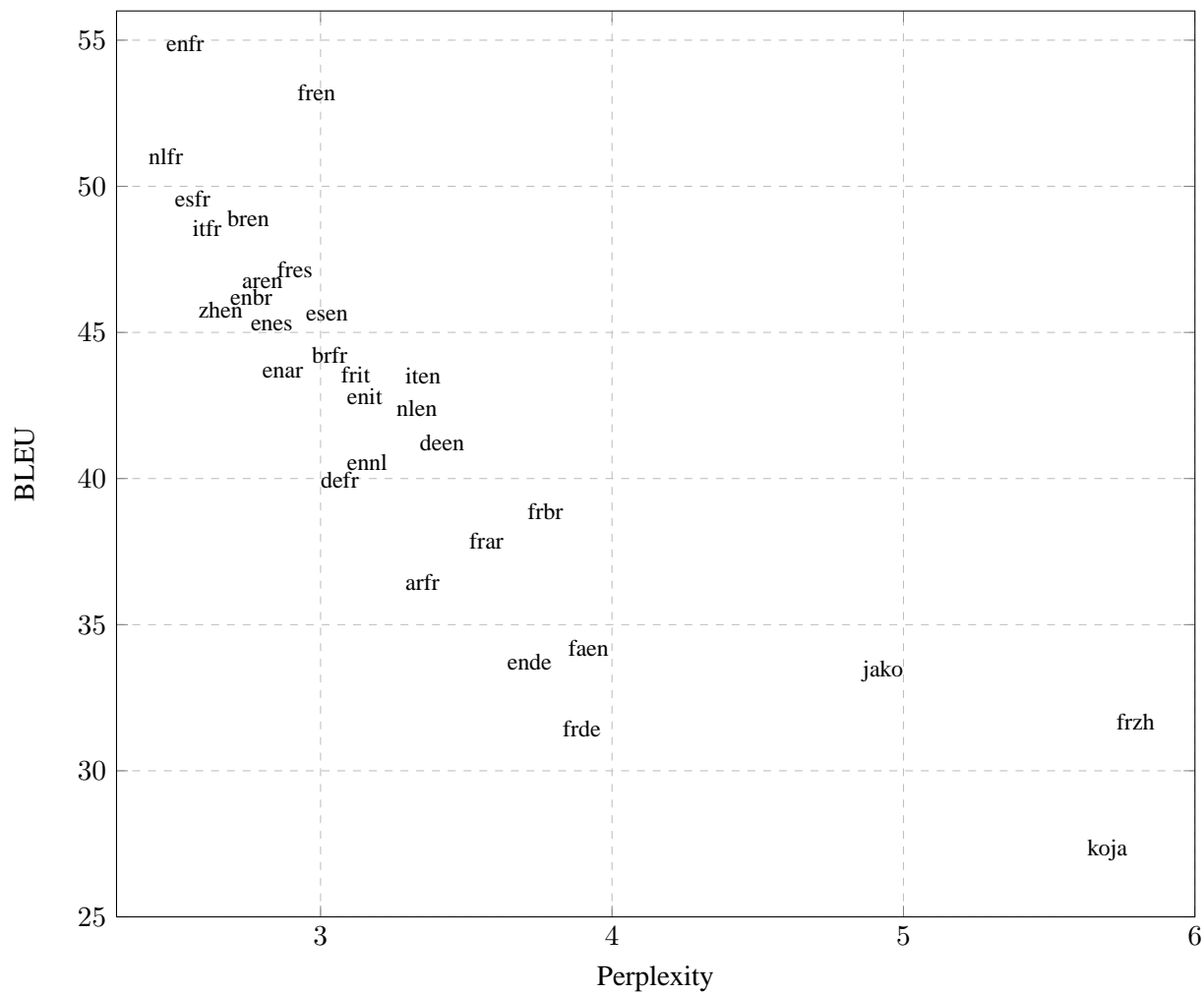

Figure \ref{fig:bleuppl} plots automatic accuracy results, BLEU, and Perplexities for all language pairs. 
It is remarkable the high correlation between perplexity and BLEU scores, showing that language pairs with lower perplexity yield higher BLEU scores. 
Note also that different amounts of training data were used for each system (see Table \ref{tab:corpora}). 
BLEU scores were calculated over an internal test set. 

From the beginning of this report we have used ``internal'' validation and test sets, what makes it difficult to compare the performance of our systems to other research engines. However, we must keep in mind that our goal is to account for improvements in our production systems. We focus on human evaluations rather than on any automatic evaluation score.

\subsection{Human Ranking Evaluation}
\label{ssec:eval_human}
To evaluate translation outputs and compare with human translation, we have defined the following protocol.
\begin{enumerate}\itemsep0em
\item For each language pair, 100 sentences ``in domain'' ({\small *}) are collected,
\item These sentences are sent to human translation ({\small **}), and translated with candidate model and using available online translation services ({\small ***}).
\item Without notification of the mix of human and machine translation, a team of 3 professional translators or linguists fluent in both source and target languages is then asked to rank 3 random outputs for each sentence based on their preference as translation. Preference includes accuracy as a priority, but also fluency of the generated translation. They have the choice to give them 3 different ranks, or can also decide to give 2 or the 3 of them the same rank, if they cannot decide.
\end{enumerate}

\noindent ({\small *}) for Generic domain, sentences from recent news article were selected online, for Technical (IT) sentences part of translation memory defined in section \ref{ssec:adaptation} were kept apart from the training.\\
({\small **}) for human translation, we did use translation agency ({\it human-pro}), and online collaborative translation platform ({\it human-casual}).\\
({\small ***}) we used Naver Translator\footnote{\url{http://translate.naver.com}} (Naver), Google Translate\footnote{\url{http://translate.google.com}} (Google) and Bing Translator\footnote{\url{http://translator.bing}} (Bing).\\

For this first release, we experimented on the evaluation protocol for 2 different extremely different categories of language pairs. On one hand, English$\leftrightarrow$French which is probably the most studied language pairs for MT and for which resources are very large: \cite{GNMT} mention about 36M sentence pairs used in their NMT training and the equivalent PBMT is completed by a web-scale target side language models\footnote{In 2007, Google already mentions using 2 trillion words in their language models for machine translation \cite{brants2007large}.}. 
Also, as English is a low inflected language, the current phrase-based technology for target language English is more competitive due to the relative simplicity of the generation and weight of gigantic language models. 

On the other hand, English$\leftrightarrow$Korean is one of the toughest language pair due to the far distance between English and Korean language, the small availability of training corpus, and the rich agglutinative morphology of Korean. For a real comparison, we ran evaluation against Naver Translation service from English into Korean, where Naver is the main South Korean search engine.

Tables \ref{tab:systemeval} and \ref{tab:evaluations} describe the different evaluations and their results.

\begin{table*}[p]
\begin{center}
\begin{tabular}[center]{|l|l|l|l|}
\hline
\textbf{Language Pair} & \textbf{Domain} ({\small *}) & \textbf{Human Translation} ({\small **}) & \textbf{Online Translation} ({\small ***}) \\\hline\hline
English $\mapsto$ French & Generic & {\it human-pro}, {\it human-casual} & Bing, Google \\ \hline
French $\mapsto$ English & Generic & {\it human-pro}, {\it human-casual} & Bing, Google \\ \hline
English $\mapsto$ Korean & News & {\it human-pro} & Naver, Google, Bing \\ \hline
English $\mapsto$ Korean & Technical (IT) & {\it human-pro} & {\it N/A} \\ \hline
\end{tabular}
\caption{\label{tab:systemeval} Performed evaluations}
\end{center}
\end{table*}

\begin{table*}[p]
\begin{center}
\begin{tabular}[center]{|l|c|c|c|c|c|c|}
\hline
 & \textbf{Human-pro} & \textbf{Human-casual} & \textbf{Bing} & \textbf{Google} & \textbf{Naver} & \textbf{Systran V8} \\ \hline
English $\mapsto$ French & {\color{red}-64.2} & {\color{red}-18.5} & {\color{blue}+48.7} & {\color{blue}+10.4} & & {\color{blue}+17.3} \\ \hline
French $\mapsto$ English & {\color{red}-56.8} & {\color{blue}+5.5} & {\color{red}-23.1} & {\color{red}-8.4} & & {\color{blue}+5} \\ \hline
English $\mapsto$ Korean & {\color{red}-15.4} & & {\color{blue}+35.5} & {\color{blue}+33.7} & {\color{blue}+31.4} & {\color{blue}+13.2} \\ \hline
English $\mapsto$ Korean (IT) & {\color{blue}+30.3} & & & & & \\ \hline
\end{tabular}
\caption{\label{tab:evaluations} This table shows relative preference of SYSTRAN NMT compared to other outputs calculated this way: for each triplet where output $A$ and $B$ were compared, we note $\textit{pref}_{A>B}$ the number of times where $A$ was strictly preferred to $B$, and $E_A$ the total number of triplet including output $A$. For each output, $E$, the number in the table is $\textit{compar}(\text{SNMT},E)=(\textit{pref}_{\text{SNMT}>E}-\textit{pref}_{E>\text{SNMT}})/E_\text{SNMT}$. $\textit{compar}(\text{SNMT},E)$ is a percent value in the range $[-1;1]$}
\end{center}
\end{table*}

Several interesting outcomes:
\begin{itemize}\itemsep0em
  \item Our vanilla English $\mapsto$ French model outperforms existing online engines and our best of breed technology.
  \item For French $\mapsto$ English, if the model slightly outperforms (human-casual) and our best of breed technology, it stays behind Google Translate, and more significantly behind Microsoft Translator.
  \item The generic English $\mapsto$ Korean model shows closer results with human translation and outperforms clearly existing online engines.
  \item The ``in-domain'' specialized model surprisingly outperforms the reference human translation.
\end{itemize}

We are aware that far more evaluations are necessary and we will be launching a larger evaluation plan for our next release. Informally, we do observe that the biggest performance jump are observed on complicated language pairs, like English-Korean or Japanese-English showing that NMT engines are better able to handle major structure differences, but also on the languages with lower resources like Farsi-English demonstrating that NMT is able to learn better with less, and we will explore this even more.

Finally, we are also launching in parallel, a real-life beta-testing program with our customers so that we can also obtain formal feedback from their use-case and related to specialized models.

\subsection{Qualitative Evaluation}
\label{ssec:eval_qualitative}
In the table \ref{tab:qualitative_eval}, we report the result of error analysis for NMT, SMT and RBMT for English-French language pair. This evaluation confirms the translation ranking performed in the previous but also exhibits some interesting facts:
\begin{itemize}
  \item The most salient error comes from missing words or parts of sentence. It is interesting to see though that half of these ``omissions'' are considered okay by the reviewers and were most of the time not considered as errors - it indeed shows the ability of the system not only to translate but to summarize and get to the point as we would expect from human translation. Of course, we need to fix the cases, where the ``omissions'' are not okay.
  \item Another finding is that the engine is badly managing quotes, and we will make sure to specifically teach that in our next release. Other low-hanging fruit are the case generation, which seems sometimes to get off-track, and the handling of Named Entity that we have already introduced in the system but not connected for the release.
  \item On the positive side, we observe that NMT is drastically improving fluency, reduces slightly meaning selection errors, and handle better morphology although it does not have yet any specific access to morphology (like sub-word embedding). Regarding meaning selection errors, we will focussing on teaching more expressions to the system which is still a major structural weakness compared to PBMT engines.
\end{itemize}

\begin{table*}[p]
\begin{center}
\begin{tabular}[t]{|ll|c|c|c|l|}
\hline
\multicolumn{2}{|l|}{\textbf{Category}} &  \textbf{NMT} & \textbf{RB} & \textbf{SMT} & \textbf{Example} \\
\hline
\multicolumn{6}{|l|}{Entity} \\ \cline{3-5}
\cdashline{2-6}[.4pt/1pt] & Major      &  7   & 5  & 0 & {\small \textit{Galaxy Note 7} $\mapsto$ \textit{\color{red}{note 7 de Galaxy}} vs. \textit{\color{blue}{Galaxy Note 7}}} \\
\cdashline{2-6}[.4pt/1pt] & Format     &  3   & 1  & 1 & (number localization): {\small \textit{\$2.66} $\mapsto$ \textit{2\color{red}{.}66 \$} vs. \textit{2\color{blue}{,}66 \$}} \\
\hline
\multicolumn{6}{|l|}{Morphology} \\ \cline{3-5}
\cdashline{2-6}[.4pt/1pt] & Minor - Local &  3   & 2  & 3 & (tense choice): {\small \textit{accused} $\mapsto$ \textit{\color{red}{accusait}} vs. \textit{\color{blue}{a accusé}}} \\
\cdashline{2-6}[.4pt/1pt] & Minor - Sentence Level & 3    & 3  & 5 & {\shortstack[l]{\small \textit{\underline{the} president [...], \underline{she} emphasized}\\\hspace{0.5cm}\small$\mapsto$ \textit{\underline{\color{red}{la}} pr\'esident [...], \underline{elle} a souligné} vs. \textit{\underline{\color{blue}{la}} [...], \underline{elle}}}}\\
\cdashline{2-6}[.4pt/1pt] & Major  &  3   & 4  & 6  & {\small \textit{he scanned} $\mapsto$ \textit{il \color{red}{scann\underline{\'e}}} vs. \textit{il \color{blue}{scann\underline{ait}}}}\\
\hline
\multicolumn{6}{|l|}{Meaning Selection} \\ \cline{3-5}
\cdashline{2-6}[.4pt/1pt] & Minor       &  9   & 17 & 7  & \small \textit{game} $\mapsto$ \textit{\color{red}{jeu}} vs. \textit{\color{blue}{match}} \\
\cdashline{2-6}[.4pt/1pt] & Major - Prep Choice         & 4    & 9  & 10 & \shortstack[l]{\small \textit{[{... facing war crimes charges}] \underline{over} [{its bombardment of ...}]}\\\hspace{0.5cm}\small$\mapsto$ \textit{\color{red}{contre}} vs. \textit{\color{blue}{pour}}}\\
\cdashline{2-6}[.4pt/1pt] & Major - Expression          & 3    & 7  & 1 & \shortstack[l]{\small \textit{[two] \underline{counts of murder}}\\\hspace{0.5cm}\small$\mapsto$ \textit{\color{red}{chefs de meutre}} vs. \textit{\color{blue}{chefs d'accusation de meutre}}}\\
\cdashline{2-6}[.4pt/1pt] & Major - Not Translated      & 5    & 1  & 4 & {\small \textit{he scanned} $\mapsto$ \textit{il \color{red}{scanned}} vs. \textit{il \color{blue}{a scanné}}}\\
\cdashline{2-6}[.4pt/1pt] & Major - Contextual Meaning  & 14   & 39 & 14 & \shortstack[l]{\small \textit{33 senior Republicans}\\\hspace{0.5cm}\small$\mapsto$ \textit{33 r\'epublicains \color{red}{sup\'erieurs}} vs. \textit{33 \color{blue}{t\'enors} r\'epublicains}}\\
\hline
\multicolumn{6}{|l|}{Word Ordering and Fluency} \\ \cline{3-5}
\cdashline{2-6}[.4pt/1pt] & Minor     & 2    & 28 & 15 & \shortstack[l]{(determiner):\small \textit{[without] a [specific destination in mind]}\\\hspace{0.5cm}\small$\mapsto$ \textit{sans \color{red}{une} destination [...]} vs. \textit{sans destination [...]}}\\
\cdashline{2-6}[.4pt/1pt] & Major     & 3    & 16 & 15 & \shortstack[l]{(word ordering):\small \textit{in the Sept. 26 deaths}\\\hspace{0.5cm}\small$\mapsto$ \textit{dans les morts \color{red}{septembre de 26}}\\\hspace{1cm}\small vs. \textit{dans les morts du \color{blue}{26 septembre}}}\\
\hline
\multicolumn{6}{|l|}{Missing or Duplicated} \\ \cline{3-5}
\cdashline{2-6}[.4pt/1pt] & Missing Minor       & 7    & 3  & 1 & \shortstack[l]{\small \textit{a week after the hurricane \underline{struck}}\\\hspace{0.5cm}\small$\mapsto$ \textit{une semaine après l'ouragan}\\\hspace{1cm}\small vs. \textit{une semaine après \textcolor{blue}{que} l'ouragan \textcolor{blue}{ait frappé}}} \\
\cdashline{2-6}[.4pt/1pt] & Missing Major       & 6    & 1  & 3 & \shortstack[l]{\small \textit{\underline{As a working oncologist,} Giordano knew [...]}\\\hspace{0.5cm}\small$\mapsto$ \textit{Giordano savait}\\\hspace{1cm}\small vs. \textit{\textcolor{blue}{En tant qu'oncologue en fonction,} Giordano savait}}\\
\cdashline{2-6}[.4pt/1pt] & Duplicated Major         & 2    & 2  & 1 & \shortstack[l]{\small \textit{for the \underline{Republican} presidential nominee}\\\hspace{0.5cm}\small$\mapsto$ \textit{au candidat républicain \textcolor{red}{républicain}}\\\hspace{1cm}\small vs. \textit{au candidat républicain}}\\
\hline
\multicolumn{6}{|l|}{Misc. (Minor)} \\ \cline{3-5}
\cdashline{2-6}[.4pt/1pt] & Quotes, Punctuations& 2    & 0  & 0 & (misplaced quotes)\\
\cdashline{2-6}[.4pt/1pt] & Case                & 6    & 0  & 2 & \shortstack[l]{\small \textit{[...] will be affected by Brexit}\\\hspace{0.5cm}\small$\mapsto$ \textit{[...] \textcolor{red}{S}era touchée \textcolor{red}{P}ar le \textcolor{red}{b}rexit}\\\hspace{1cm}\small vs. \textit{[...] \textcolor{blue}{s}era touchée \textcolor{blue}{p}ar le \textcolor{blue}{B}rexit}}\\
\hline \hline
\multicolumn{5}{|l}{\textbf{Total}} \\
\cline{3-5} & \textbf{Major} & 47  & 84 & 54  \\
\cline{3-5} & \textbf{Minor} & 36  & 55 & 35  \\
\cline{3-5} & \textbf{Minor \& Major} & \textbf{83}  & \textbf{139} & \textbf{89}  \\
\cline{1-5}
\end{tabular}
\caption{\label{tab:qualitative_eval} Human error analysis done for 50 sentences of the corpus defined in the section \ref{ssec:eval_human} for English-French on NMT, SMT (Google) and RBMT outputs. Error categories are: - issue with entity handling (Entity), issue with Morphology either local or reflecting sentence level missing agreements, issue with Meaning Selection splitted into different sub-categories, - issue with Word Ordering or Fluency (wrong or missing determiner for instance), - missing or duplicated words. Errors are either Minor when reader could still understand the sentence without access to the source, otherwise is considered as Major. Erroneous words are counted in only one category even if several problems add-up - for instance ordering and meaning selection.}
\end{center}
\end{table*}

\section{Practical Issues}
\label{sec:practical}

Translation results from an NMT system, at first glance, is incredibly fluent that your are diverted from its downsides.
Over the course of the training and during our internal evaluation, we found out multiple practical issues worth sharing:
\begin{enumerate}
  \item translating very long sentences
  \item translating user input such as a short word or the title of a news article
  \item cleaning the corpus
  \item alignment
\end{enumerate}

\begin{figure*}[p]
\includegraphics[width=15cm]{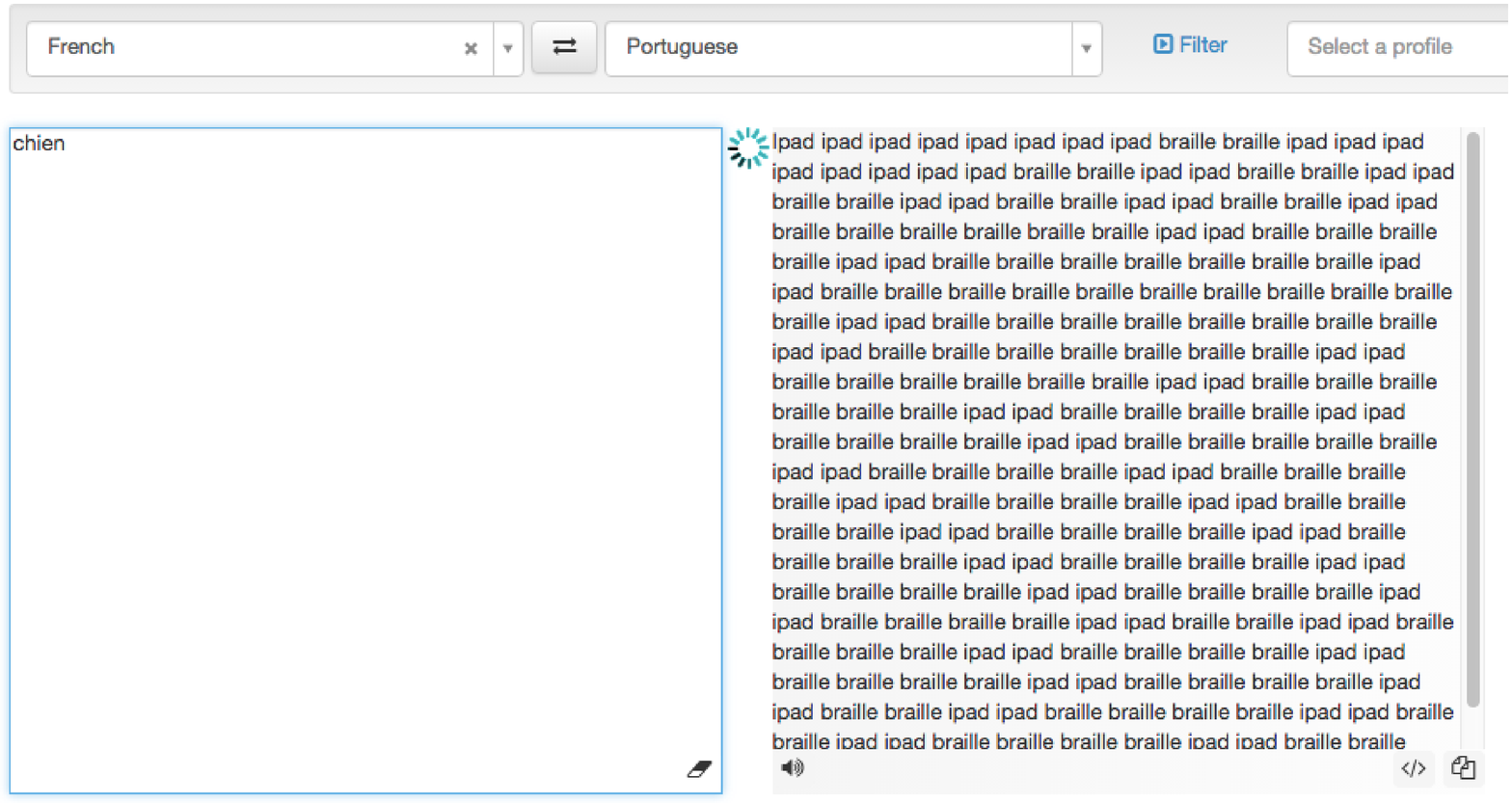}
\caption{\label{fig:wordTranslator} Effect of translation of a single word through a model not trained for that.}
\end{figure*}

NMT is greatly impacted by the train data, on which NMT learns how to generate accurate and fluent translations holistically.
Because the maximal length of a training instance was limited to a certain length during the training of our models, NMT models are puzzled by sentences that exceed this length, not having encountered such a training data. Hard splitting of longer sentences has some side-effects since the model consider both parts as full sentence. As a consequence, whatever is the limit we set for sentence length, we do also need to teach the neural network how to handle longer sentences. For that, we are exploring several options including using separate model based on source/target alignment to find optimal breaking point, and introduce special \texttt{<TO BE CONTINUED>} and \texttt{<CONTINUING>} tokens. Likewise, very short phrases and incomplete or fragmental sentences were not included in our training data,
and consequently NMT systems fail to correctly translate such input texts (e.g. Figure \ref{fig:wordTranslator}). Here also, to enable this, we do simply need to teach the model to handle such input by injecting additional synthetic data.

Also, our training corpus includes a number of resources that are known to contain many noisy data. While NMT seems more robust than other technologies for handling noise, we can still perceive noise effect in translation - in particular for recurring noise. An example is in English-to-Korean, where we see the model trying to systematically convert amount currency in addition to the translation. As demonstrated in Section \ref{ssec:distillation}, preparing the \emph{right kind of input} to NMT seems to result in more efficient and accurate systems, and such a procedure should also be directly applied to the training data more aggressively.

Finally,let us note that source/target alignment is a must for our users, but this information is missing from NMT output due to soft alignment. To hide this issue from the end-users, multiple alignment heuristic are showing tradictional target-source alignment.

\section{Further Work}
\label{sec:furtherwork}

In this section we outline further experiments currently being conducted. First we extend NMT decoding with the ability to make use of multiple models. Both external models, particularly an $n$-gram language model, as well as decoding with multiple networks (ensemble). We also work on using external word embeddings, and on modelling unknown words within the network.

\subsection{Extending Decoding}
\subsubsection{Additional LM}
\label{sssec:additionallm}

As proposed by ~\cite{monolingualcorporaNMT}, we conduct experiments to integrate an $n$-gram language model estimated over a large dataset on our Neural MT system. We followed a shallow fusion integration, similar to how language models are used in a standard phrase-based MT decoder. 

In the context of beam search decoding in NMT, at each time step $t$, candidate words $x$ are hypothesized and assigned a score according to the neural network, $p_{NMT}(x)$. Sorted according to their respective scores, the $K$-best candidates, are reranked using the score assigned by the language model, $p_{LM}(x)$. The resulting probability of each candidate is obtained by the weighted sum of each log-probability $log\ p(x) = log\ p_{LM}(x) + \beta\ log\ p_{NMT}(x)$. Where $\beta$ is a hyper-parameter that needs to be tuned. 

This technique is specially useful for handling out-of-vocabulary words (OOV). Deep networks are technically constrained to work with limited vocabulary sets (in our experiments we use target vocabularies of $60$k words), hence suffering from important OOV problems. In contrast, $n$-gram language models can be learned for very large vocabularies.

Initial results show the suitability of the shallow integration technique to select the appropriate OOV candidate out of a dictionary list (external resource). The probability obtained from a language model is the unique modeling alternative for those word candidates for which the neural network produces no score.

\subsubsection{Ensemble Decoding}
\label{sssec:ensemble}

Ensemble decoding has been verified as a practical technique to further improve the performance
compared to a single Encoder-Decoder model \cite{Sennrich2016,GNMT,zhou2016deep}.
The improvement comes from the diversity of prediction from different neural network models,
which are learned by random initialization seeds and shuffling of examples during training,
or different optimization methods towards the development set\cite{cho2015using}.
As a consequence, 3-8 isolated models will be trained and ensembled together,
considering the cost of memory and training speed. Also, \cite{WIPO} provides some
methods to accelerate the training by choosing different checkpoints as the final models.

We implement ensemble decoding by averaging the output probabilities for each estimation of
target word \(x\) with the formula:

\begin{center}
\(
p_{x}^{ens}
=\frac{1}{M}
\sum_{m=1}^{M}
p_{x}^{m}
\)
\end{center}

\noindent wherein, \(p_{x}^{m}\) represents probabilities of each \(x\), and \(M\) is the number of neural models.

\subsection{Extending word embeddings}
\label{sssec:wordembeddings}

Although NMT technology has recently accomplished a major breakthrough in Machine Translation field, it still remains constrained due to the limited vocabulary size and to the use of bilingual training data. In order to reduce the negative impact of both phenomena, experiments are currently being held on using external word embedding weights.

Those external word embeddings are not learned by the NMT network from bilingual data only, but by an external model (e.g. word2vec \cite{DBLP:journals/corr/abs-1301-3781}). They can therefore be estimated from larger monolingual corpora, incorporating data from different domains.

Another advantage lies in the fact that, since external word embedding weights are not modified during NMT training, it is possible to use a different vocabulary for this fixed part of the input during the application or re-training of the model (provided that the weights for the words in new vocabulary come from the same embedding space as the original ones). This may allow a more efficient adaptation to the data coming from a different domain with a different vocabulary.


\subsection{Unknown word handling}
\label{ssec:positional softmax}

When an \emph{unknown word} is generated in the target output sentence, a general encoder-decoder with attentional mechanism utilizes heuristics based on attention weights such that the source word with the most attention is either directly copied as-is or looked up in a dictionary.

In the recent literature \cite{gu-EtAl:2016:P16-1,gulcehre-EtAl:2016:P16-1}, researchers have attempted to directly model the unknown word handling within the attention and decoder networks.
Having the model learn to take control of both decoding and unknown word handling will result in the most optimized way to address the single unknown word replacement problem, and we are implementing and evaluating the previous approaches within our framework.

\section{Conclusion}
\label{sec:conclusions}
Neural MT has progressed at a very impressive rate, and it has proven itself to be competitive against online systems trained on train data whose size is several orders of magnitude larger.
There is no doubt that Neural MT is definitely a technology that will continue to have a great impact on academia and industry.
However, at its current status, it is not without limitations;
on language pairs that have abundant amount of monolingual and bilingual train data, phrase-based MT still perform better than Neural MT, because Neural MT is still limited on the vocabulary size and deficient utilization of monolingual data.

Neural MT is not an one-size-fits-all technology such that one general configuration of the model universally works on any language pairs.
For example, subword tokenization such as BPE provides an easy way out of the limited vocabulary problem, but we have discovered that it is not always the best choice for all language pairs.
Attention mechanism is still not at the satisfactory status and it needs to be more accurate for better controlling the translation output and for better user interactions.

For upcoming releases, we have begun to making even more experiments with injection of various linguistic knowledges, at which SYSTRAN possesses the foremost expertise.
We will also apply our engineering know-hows to conquer the practical issues of NMT one by one.


\section*{Acknowledgments}

We would like to thank Yoon Kim, Prof. Alexander Rush and the rest of members of the Harvard NLP group for their support with the open-source code, their pro-active advices and their valuable insights on the extensions.

We are also thankful to CrossLang and Homin Kwon for their thorough and meaningful definition of the evaluation protocol, and their evaluation team as well as Inah Hong, Weimin Jiang and SunHyung Lee for their work.

\bibliographystyle{acl2016}
\bibliography{acl2016}
\onecolumn
\newpage
\appendix
\section{Remarkable Results}

In this section, we highlight a serie of ``remarkable'' translations (positively remarkable and also few negatively outstanding) that we found out during evaluation for a variety of languages.

\begin{table}[h]
\small
\begin{center}
\begin{tabular}[c]{|l|p{4.5cm}|p{4.5cm}|p{4.5cm}|}
\hline
\shortstack{\textbf{Language}\\\textbf{Pair}}  & \textbf{Source} & \textbf{NMT} & \textbf{Google Translate (2016/10/15)} \\
\hline
en$\mapsto$fr
  & The report also said that although three-quarters of Chinese say their country is playing a larger role in the world than it did a decade ago, most want their government to focus on domestic issues rather than helping other nations.
  & {\color{blue}Le rapport dit aussi que, bien que les trois quarts des Chinois disent que leur pays joue un rôle plus important dans le monde qu'il \underline{ne} l'a fait il y a dix ans, la plupart veulent que leur gouvernement se concentre sur les questions nationales plutôt que d'aider d'autres nations.}
  & Le rapport indique également que, bien que les trois quarts des Chinois disent que leur pays joue un rôle plus important dans le monde qu'il y a dix ans, la plupart veulent que leur gouvernement de se concentrer sur les questions intérieures plutôt que d'aider les autres nations. \\
fr$\mapsto$en
  & Fenêtre à rue se fermait mal, très très bruyant et mal dormi. Petit déjeuner banal.
  & {\color{red}The room was very clean and the room was very clean.}
  & street window will not close, very very noisy and slept badly. mundane breakfast. \\
en$\mapsto$ko
  & Forty Republican senators and congressmen have revoked their support for the Republican presidential nominee — with nearly 30 of them calling on him to quit the race altogether in recent days.
  & {\color{blue}40 명의 공화당 상원의원들과 하원의원들이 공화당 대통령 후보자에 대한 그들의 지지를 철회했고, 거의 30 명의 사람들이 최근 며칠 동안 그 경주를 그만두도록 그에게 요청했습니다.}
  & (From Naver NMT\footnote{http://labspace.naver.com/nmt/ - on 2016/10/17}): 40여명의 공화당 의원들과 하원 의원들은 거의 30명의 공화당 대통령 후보들을 철회했고, 그들 중 거의 30명이 최근에 그를 은퇴 직전에 그만두게 할 것을 촉구했다. \\
en$\mapsto$it
  & all this leads to feelings of insecurity and conflict thus reinforcing negative spirals.
  & {\color{blue}tutto ciò porta a sentimenti di insicurezza e di conflitto, rafforzando così la spirale negativa.}
  & tutto ciò genera elementi di insicurezza e di conflitto rinforzando quindi spirali negative. \\
it$\mapsto$fr
  & Mi sembra che ciò non abbia nulla a che vedere con una penuria alimentare: si tratta di una vera carestia, di una vera e propria catastrofe che non è naturale ma strutturale, provocata da un regime folle, criminale, psicopatico, peggiore dei peggiori regimi esistiti nell'Unione Sovietica e nella Romania di Ceaucescu.
  & {\color{blue}Il me semble que cela n'a rien à voir avec une pénurie alimentaire: il s'agit d'une véritable famine, d'une véritable catastrophe qui n'est pas naturelle mais structurelle, provoquée par un régime insensé, criminel, psychopathe, pire que les pires régimes de l'union soviétique et de la Roumanie de Ceaucescu.}
  & Il me semble que cela n'a rien à voir avec une pénurie alimentaire: il est une véritable famine, une catastrophe qui est pas naturel, mais une structure, provoquée par un régime criminel fou, psychopathe, le pire des pires régimes existait l'Union soviétique et en Roumanie de Ceaucescu. \\
it$\mapsto$fr
  & Ebbene io sono in totale disaccordo: lo Stato non deve passare in secondo piano; deve invece intervenire di più e garantire una maggiore vigilanza in ambito economico.
  & {\color{red}je suis tout à fait d'accord}: l'état ne doit pas passer sous silence; il doit plutôt agir et assurer une plus grande vigilance dans le domaine économique.
  & Eh bien, je suis en désaccord total: l'Etat ne doit pas prendre un siège arrière; Il doit plutôt agir plus et assurer une plus grande supervision dans l'économie. \\
es$\mapsto$en
  & ¿Es posible un Bangladesh más amable con los niños?
  & {\color{orange}Is Bangladesh more friendly with children?}
  & Can a kinder Bangladesh with children? \\
en$\mapsto$de
  & On the other hand, businesses are constantly complaining about the shortage of specialist staff, using this supposed shortage as an excuse to use cheap labour.
  & {\color{blue}Andererseits beschweren sich Unternehmen ständig über den Mangel an Fachpersonal, indem sie diese angebliche Knappheit als Vorwand benutzen, billige Arbeitskräfte einzusetzen.}
  & Auf der anderen Seite beschweren sich Unternehmen ständig über den Mangel an Fachkräften, wobei diese angebliche Mangelhaftigkeit als Ausrede für billige Arbeitskräfte verwendet wird. \\
zh$\mapsto$en
  & \begin{CJK*}{UTF8}{gbsn} 朝闻天下,现在又到了我们的互动时间啦. \end{CJK*}
  & {\color{blue}Morning News, now it's our interactive time.}
  & (From Google GNMT): North Korea heard the world, and now to our interaction time friends. \\
\hline
\end{tabular}
\end{center}
\end{table}

\begin{table}[h]
\small
\begin{center}
\begin{tabular}[c]{|l|p{4.5cm}|p{4.5cm}|p{4.5cm}|}
\hline
\shortstack{\textbf{Language}\\\textbf{Pair}}  & \textbf{Source} & \textbf{NMT} & \textbf{Google Translate (2016/10/15)} \\
\hline

br$\mapsto$en
  & Face às decisões de Nice, tínhamos de ter em conta esta situação e de adotar uma metodologia que contemplasse as transformações necessárias.
  & {\color{blue}Faced with the Nice decisions, we had to take this situation into account and adopt a methodology that would address the necessary changes.}
  & Nice view of the decisions we had to take account of this situation and adopt a methodology that took into consideration the necessary changes. \\
fr$\mapsto$br
  & Une information présentée au moment opportun signifie la transparence, laquelle crée la confiance et évite à l'entreprise de subir des pertes.
  & {\color{blue}A informação apresentada no momento oportuno significa transparência, que cria confiança e evita que a empresa sofra perdas.}
  & Informação apresentada em uma transparência meio oportuna, que cria confiança e evita a empresa a sofrer perdas. \\

\hline
\end{tabular}
\end{center}
\end{table}

\newpage
\section{Online System Parameters}
\label{sec:parameters}

All systems were trained with $4$ LSTM layers, size of word embedding vectors was $500$, dropout was set to $0.3$ and we used bidirectional RNN (BRNN). Column Guided Alignment indicates wether the network was trained with guided alignments and on which epoch the feature was stopped.

\begin{table}[h]
\small
\begin{center}
\begin{tabular}{|c||c|c|c|c|c|c|c|c|c|c|}
\hline
 & Tokenization & \shortstack{RNN\\size} & \shortstack{Optimal\\Epoch} & \shortstack{Guided\\Alignment} & \shortstack{NER\\aware} & Special \\
\hline
zh$\mapsto$en & word boundary-generic & 800 & 12 &  epoch 4 & yes & \\ 
en$\mapsto$it & generic & 800 & 16 &  epoch 4 & yes & \\ 
it$\mapsto$en & generic & 800 & 16 &  epoch 4 & yes & \\ 
en$\mapsto$ar & generic-crf  & 800 & 15 &  epoch 4 & no & \\ 
ar$\mapsto$en &  crf-generic & 800 & 15 &  epoch 4 & no & \\ 
en$\mapsto$es & generic  & 800 & 18 &  epoch 4 & yes & \\ 
es$\mapsto$en & generic  & 800 & 18 &  epoch 4 & yes & \\ 
en$\mapsto$de & generic-compound splitting  & 800 & 17 &  epoch 4 & yes & \\ 
de$\mapsto$en & compound splitting-generic  & 800 & 18 &  epoch 4 & yes & \\ 
en$\mapsto$nl & generic & 800 & 17 &  epoch 4 & no & \\ 
nl$\mapsto$en & generic  & 800 & 14 &  epoch 4 & no & \\ 
en$\mapsto$fr & generic  & 800 & 18 &  epoch 4 & yes & double corpora ($2\ x\ 5$M)\\ 
fr$\mapsto$en & generic & 800 & 17 &  epoch 4 & yes & \\ 
ja$\mapsto$en & bpe      & 800 & 11 & no & no & \\ 
\hline
fr$\mapsto$br & generic & 800 & 18 &  epoch 4 & yes & \\ 
br$\mapsto$fr & generic & 800 & 18 &  epoch 4 & yes & \\ 
en$\mapsto$pt & generic & 800 & 18 &  epoch 4 & yes & \\ 
br$\mapsto$en & generic & 800 & 18 &  epoch 4 & yes & \\ 
fr$\mapsto$it & generic & 800 & 18 &  epoch 4 & yes & \\ 
it$\mapsto$fr & generic & 800 & 18 &  epoch 4 & yes & \\ 
fr$\mapsto$ar & generic-crf  & 800 & 10 &  epoch 4 & no & \\ 
ar$\mapsto$fr & crf-generic  & 800 & 15 &  epoch 4 & no & \\ 
fr$\mapsto$es & generic  & 800 & 18 &  epoch 4 & yes &\\ 
es$\mapsto$fr & generic & 800 & 18 &  epoch 4 & yes & \\ 
fr$\mapsto$de & generic-compound splitting  & 800 & 17 &  epoch 4 & no & \\ 
de$\mapsto$fr & compound splitting-generic & 800 & 16 &  epoch 4 & no & \\ 
nl$\mapsto$fr & generic & 800 & 16 &  epoch 4 & no & \\ 
fr$\mapsto$zh & generic-word segmentation  & 800 & 18 &  epoch 4 & no & \\ 
\hline
ja$\mapsto$ko & bpe & 1000 & 18 & no & no & \\ 
ko$\mapsto$ja & bpe & 1000 & 18 & no & no & \\ 
en$\mapsto$ko & bpe & 1000 & 17 & no & no & politeness\\ 
fa$\mapsto$en & basic & 800 & 18 & yes & no & \\ 
\hline
\end{tabular}
\end{center}
\caption{\label{tab:parameters} Parameters for online systems. }
\end{table}

\end{document}